\pdfoutput=1

\documentclass[11pt]{article}

\usepackage[]{emnlp2021}

\usepackage{times}
\usepackage{latexsym}
\usepackage{amsthm,amsmath,amssymb}
\usepackage{mathrsfs}
\usepackage{algorithm}  
\usepackage{algorithmicx}  
\usepackage{algpseudocode}

\usepackage[T1]{fontenc}

\usepackage[utf8]{inputenc}

\usepackage{microtype}
\usepackage{graphicx, subfigure}

\title{Asking Questions Like Educational Experts: {A}utomatically Generating Question-Answer Pairs on Real-World Examination Data}

\author{Fanyi Qu, Xin Jia, Yunfang Wu\thanks{ \; Corresponding author} \\
  Key Laboratory of Computational Linguistics (Peking University), Ministry of Education  \\
  Department of Computer Science and Technology, Peking University \\
  \texttt{\{fanyiqu, jemmryx, wuyf\}@pku.edu.cn} \\}

\begin{document}
\maketitle
\begin{abstract}
Generating high quality question-answer pairs is a hard but meaningful task. Although previous works have achieved great results on answer-aware question generation, it is difficult to apply them into practical application in the education field. This paper for the first time addresses the question-answer pair generation task on the real-world examination data, and proposes a new unified framework on RACE. To capture the important information of the input passage we first automatically generate (rather than extracting) keyphrases, thus this task is reduced to keyphrase-question-answer triplet joint generation. Accordingly, we propose a multi-agent communication model to generate and optimize the question and keyphrases iteratively, and then apply the generated question and keyphrases to guide the generation of answers. To establish a solid benchmark, we build our model on the strong generative pre-training model. Experimental results show that our model makes great breakthroughs in the question-answer pair generation task. Moreover, we make a comprehensive analysis on our model, suggesting new directions for this challenging task. 

\end{abstract}

\section{Introduction}

Question-answer pair generation (QAG) is to do question generation (QG) and answer generation (AG) simultaneously only with a given passage. The generated question-answer (Q-A) pairs can be effectively applied in numbers of tasks such as knowledge management \cite{wagner2005supporting}, FAQ document generation \cite{krishna2019generating}, and data enhancement for reading comprehension tasks (\citealp{tang2017question, liu2020asking, liu2020tell}). Particularly, high-quality Q-A pairs can facilitate the instructing process and benefit on creating educational materials for reading practice and assessment \cite{heilman2011automatic, jia2020eqg}.


Recently, much work has devoted to QG, while the QAG task is less addressed. The existing approaches on this task can be roughly grouped into two categories: the joint learning method to treat QG and answer extraction as dual tasks (\citealp{collobert2011natural,firat2016multi}); the pipeline strategy that considers answer extraction and QG as two sequential processes \cite{cui2021onestop}.

\begin{figure}[t]
\centering
\includegraphics[width=0.8\columnwidth,height=2.5in]{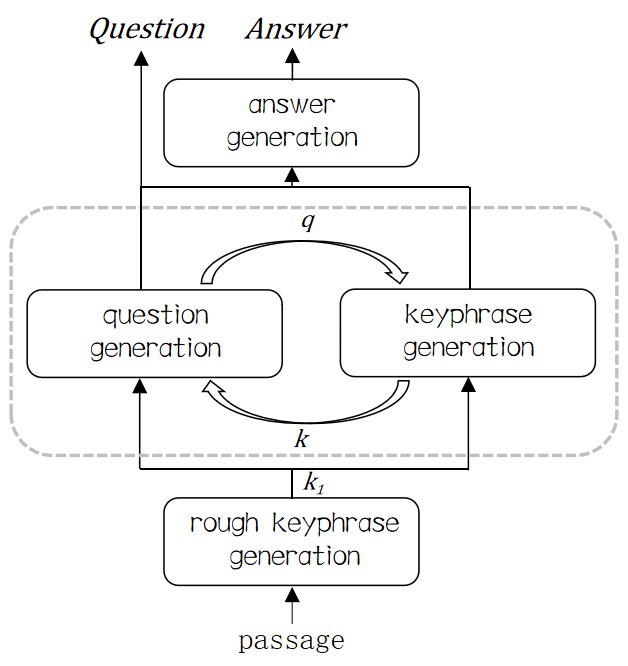}
\caption{A brief overview of our proposed framework. }
\label{fig:qk}
\end{figure}

Although some progress has been made, the QAG methods still face several challenges. Most of the existing techniques are trained and tested on the Web-extracted corpora like SQuAD \cite{rajpurkar2016squad}, MARCO \cite{nguyen2016ms} and NewsQA \cite{trischler2017newsqa}. Considering the biased and unnatural language sources of datasets, employing these techniques in educational field is difficult. Moreover, most of the previous works regard the answer text as a continuous span in the passage and directly obtain the answer through an extractive method, which may not meet the demands of real world data.


To alleviate the above limitations, we propose to perform QAG on RACE \cite{lai2017race}
. RACE is a reading comprehension corpus collected from the English exams of middle and high schools in China. Compared to Web-extracted corpora, there are two notable characteristics of RACE: real-world data distribution and generative answers, which raise new challenges for the QAG task. First, the examination-type documents have more abundant information and more diverse expressions, 
putting forward higher requirements to the generation model. Second, 
the model is required to be able to summarize the whole document rather than extracting a continuous span from the input.


In this paper, we propose a new architecture to deal with this real-world QAG task, as illustrated in Figure~\ref{fig:qk}.
It consists of three parts: rough keyphrase generation agent, question-keyphrase iterative generation module and answer generation agent. 
We first generate keyphrases based on the given document, and then optimize the generated question and keyphrases with an iterative generation module. Further, with the generated keyphrases and questions as guidance, the corresponding answer is generated. To cope with the complex expressions of examination texts, we base our model on the generative pre-training model ProphetNet \cite{qi2020prophetnet}. We conduct experiments on RACE and achieve satisfactory improvement compared to the baseline models.


Our contributions are summarized as follows:

1) We are the first to perform QAG task on RACE. The proposed methods can be easily applied to real-world examination to produce reading comprehension data. 

2) We propose a new architecture to do question-answer pair joint generation, which obtains obvious performance gain over the baseline models.

3) We conduct a comprehensive analysis on the new task and our model, establishing a solid benchmark for future researches.


\section{Data Analysis}


In this section, we present a deep dive into SQuAD and RACE to explore the challenging of QAG on real-world examination data.   


\subsection{The Questions Are Difficult}
\begin{figure}[t]
\centering
\includegraphics[width=0.98\columnwidth,height=2.5in]{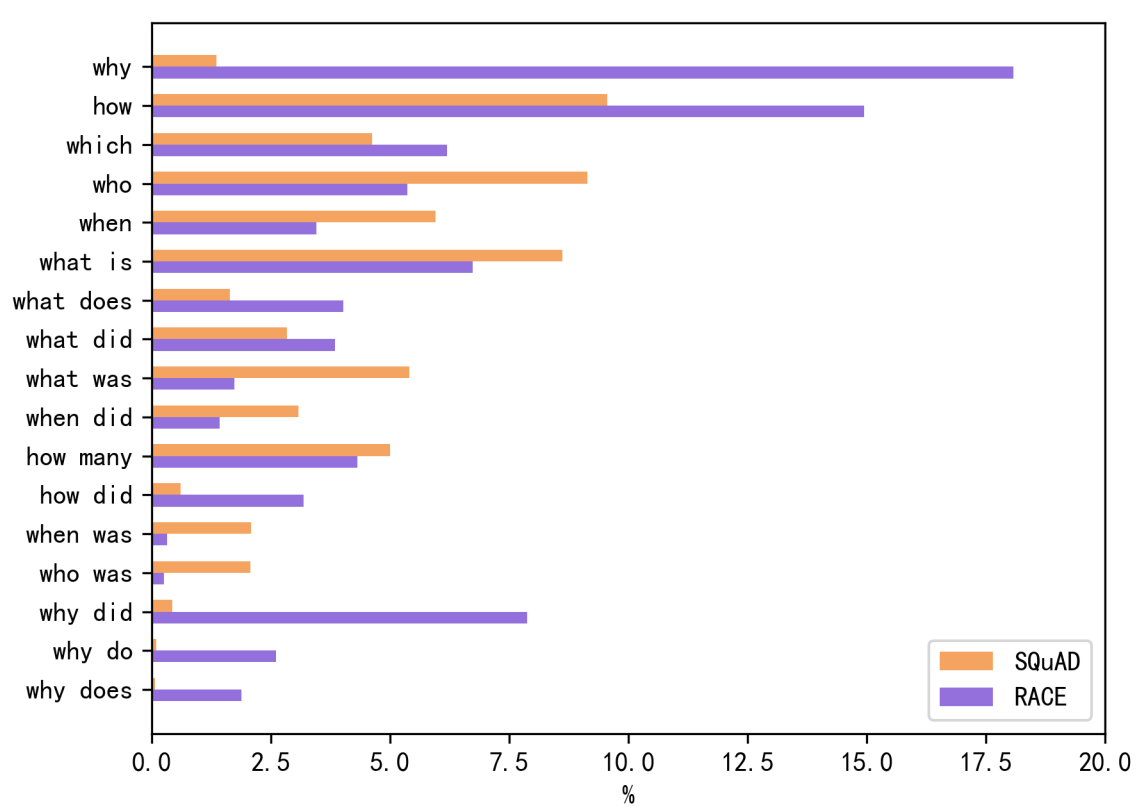}
\caption{Distribution of question types in SQuAD and RACE. }
\label{fig:ques}
\end{figure}

To get a basic sense of the question type in RACE and SQuAD, we count the proportion of the leading unigrams and bigrams that start a question for both datasets, and report the results in Figure~\ref{fig:ques}. 

Through the statistics we can reasonably conclude that questions in RACE are much more difficult than SQuAD, for ‘what’ questions (mostly detail-based) play a major role in SQuAD while RACE are more concerned with ‘why’ questions (mostly inference-based). To answer or generate inference-based questions will be more challenging than detail-based questions, since readers need to do an integration of information and conduct knowledge reasoning.


\subsection{The Answers Are Generated}
Also, we investigate the n-gram matching rate between an answer and its corresponding passage to measure the AG difficulty for both datasets. 
On SQuAD, the answers are exacted sub-spans in the passage, so the n-gram matching ratio is fixed to 100\%. However on RACE, only 68.8\% unigrams in the answer are also in the passage, and the matching ratio of bigram and trigram spans is even much lower, with 28.9\% and 14.4\% respectively. It indicates that the conventional extracting strategy of keyphrase is not appropriate for QAG task on real-world examination texts.


\section{Proposed Model}

\begin{figure*}[t]
\flushleft
\includegraphics[width=2.05\columnwidth,height=3.2in]{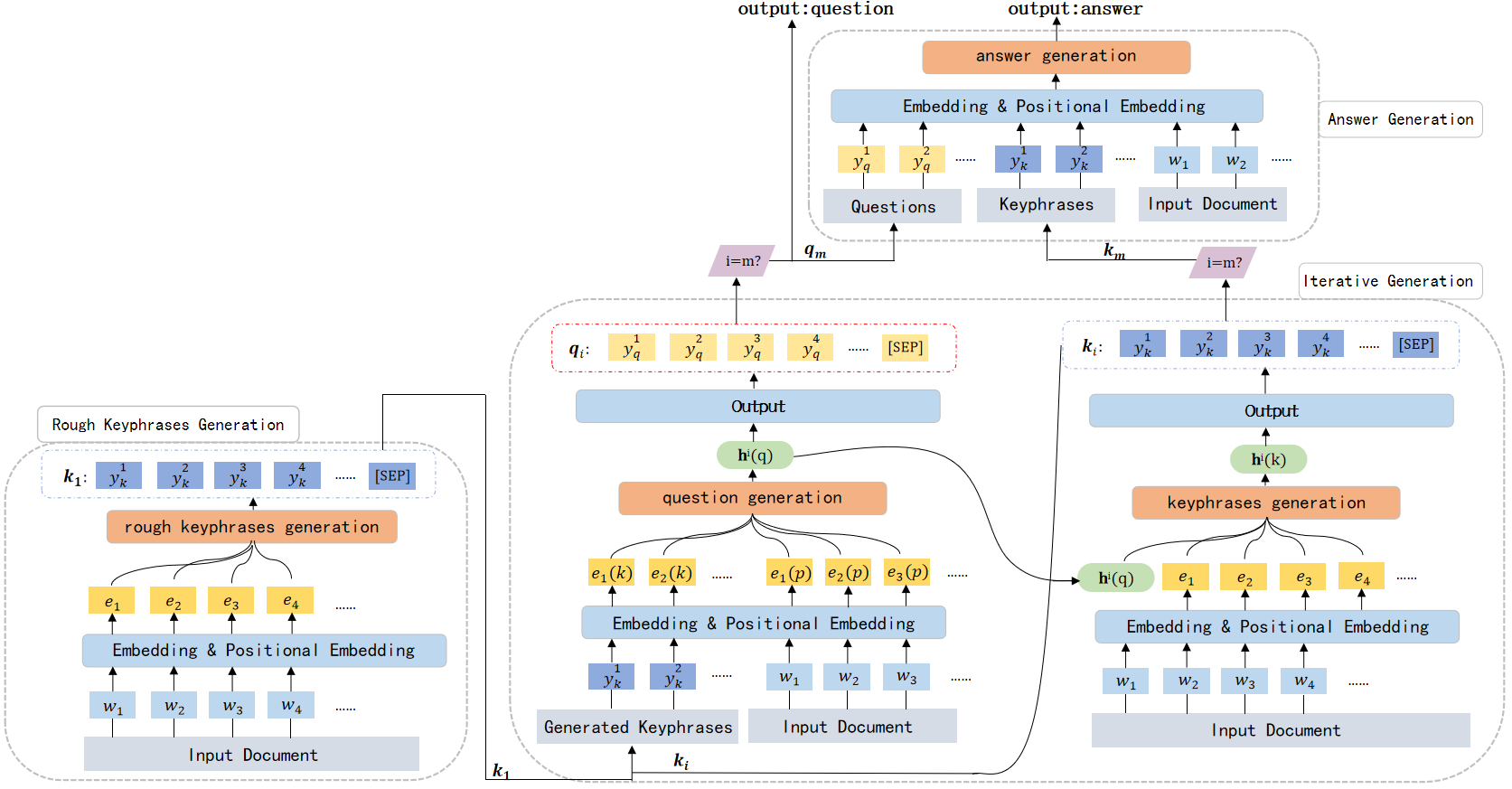}
\caption{The detailed model structure for Q-A pair joint generation} 
\label{fig:main}
\end{figure*}

\subsection{Model Overview}
In this paper, we propose a new framework for Q-A joint generation based on a generative pre-training model, and the detailed model structure is illustrated in Figure~\ref{fig:main}. The whole generation process can be split into three components:

\textbf{Step 1.} Rough keyphrase generation: generate rough keyphrases from the document, which are fed to the question generation process;

\textbf{Step 2.} Iterative question and keyphrase generation: optimize question and keyphrase iteratively with the initial input keyphrase from Step 1. 

\textbf{Step 3.} Answer generation: generate answers with the output questions and keyphrases.

To clearly describe our model, we use \emph{p} to denote the input passage,
\emph{$q_{i}$} and \emph{$k_{i}$} refer to the generated question and keyphrase in the \emph{i}-th iteration. Specially, \emph{$k_{1}$} denotes the rough keyphrase from Step 1. Let \emph{a} refer to the generated answer, and \emph{m} refer to the iterative training epochs, we can give a brief definition for \emph{q}, \emph{k}, \emph{a} as:
\begin{align}
& q_{i} = arg \mathop{max}\limits_{q}\text{P}(q|p,k_i) \\
& k_{i}=\left\{
\begin{array}{lr}
    arg \mathop{max}\limits_{k}\text{P}(k|p,q_{i-1}), & i>1\\
    arg \mathop{max}\limits_{k}\text{P}(k|p), & i=1
\end{array}
\right. \\
& a = arg \mathop{max}\limits_{a}\text{P}(a|p,k_{m},q_{m})
\end{align}

Throughout the entire training process, the objective is to minimize the negative log likelihood of the target sequence:
\begin{align}
    \mathcal{L} = -\frac{1}{T}\sum_{t=1}^{T} \text{log} P_{\theta} (y_t|y_{<t},x)
\end{align}


Our models are based on the generative pre-training model ProphetNet \cite{qi2020prophetnet}. Drawing lessons from XLNet\cite{yang2019xlnet}, ProphetNet proposes a n-stream self-attention method with a n-gram prediction pre-training task, which can be described as: 

\begin{align}
& H^{(k+1)} = \textbf{MultiHead}(H^{(k)},H^{(k)},H^{(k)}) \\
& g_{t}^{(k+1)}=\textbf{ATT}(g_{t}^{(k)}, H_{\le t}^{(k)}\oplus g_{t}^{(k)},H_{\le t } \oplus g_{t}^{(k)}) \\
& s_{t}^{(k+1)}=\textbf{ATT}(s_{t}^{(k)}, H_{\le t}^{(k)}\oplus s_{t}^{(k)},H_{\le t } \oplus s_{t}^{(k)})
\end{align}

where \emph{$H^{(k)}$} denotes the main stream self-attention, which is the same as transformer's self-attention, \emph{$g_{t}^{(k+1)}$} and \emph{$s_{t}^{(k+1)}$} denote hidden state of the 1-st and 2-nd predicting stream at time step t from the \emph{$k + 1$}-th layer, and are used to predict \emph{$y+1$} and \emph{$y+2$} respectively. $\oplus$ denotes concatenation operation. ProphetNet achieves great progress in generative tasks and obtains the best result on QG task \cite{liu2020glge}.


It is worth emphasizing that our training framework has nothing to do with the choice of the underlying model, so we can choose either normal Seq2Seq models like LSTM \cite{hochreiter1997long} and Transformer \cite{vaswani2017attention} or pre-training models like BART \cite{lewis2020bart} to replace ProphetNet.

\subsection{Two-stage Fine-tuning of Keyphrase Generation}
In this paper, we aim to generate multiple Q-A pairs with a given document as the only input. Accordingly, the vital first step is to obtain question-worthy keyphrases which provide the overall important information of the long document.  
Actually, keyphrase is an approximation to the ground-truth answer and an extraction model is often utilized in previous works. However, considering the different answer characteristics we discussed in Section 2.2, extractive methods may not work well on RACE. 
Therefore, as an alternative, we construct a ProphetNet-based keyphrase generation model to capture the key information among the document. 

There are two reasons why we choose ProphetNet for keyphrase generation. First, ProphetNet is effective enough since it is proved to be perfectly competent on several automatic generative works. More importantly, ProphetNet is employed in all the three stages of our unified model consistently, which ensures the generality and simplicity of our framework.



To further improve the quality of the generated keyphrases, we adopt a two-stage fine-tuning strategy. First, we use SQuAD as data augmentation for the first-stage training. The keyphrase generation model takes the passage as input and concatenates all the reference answers corresponding to the passage with a spacial separator as the training target. Then, we fine-tune the model meticulously on RACE dataset. Due to the characteristics of RACE, we remove stop words from the reference answers to form several separate key answer phrases, which serve as the training target in second stage training. We represent the generated result as \emph{$k_{1}$}, which is a string consisted of multiple keyphrases during inference.


\subsection{Question-Keyphrase Iterative Generation}
We propose a multi-task framework for iterative generation between question and keyphrase. Question generation is first launched taking the generated \emph{$k_{1}$} as assistance, where \emph{$k_{1}$} will be split and separately fed into the question generation model. Then the generated question is fed back to the keyphrase generation agent for optimization.

\vspace{0.2cm}
\noindent\textbf{Question Generation Agent} 

At each step, $k_i$ will be transmitted from the keyphrase generation agent to question generation process to assist the generation of $q_i$. Briefly, we concatenate $p$ and $k_i$ directly with a separator \textbf{[CLS]} as the input of the encoder:
\begin{equation}
v_{q}^{i} = \text{Enc}_{\mathcal{M}_{q}^{i}}(f([p:k_i]))    
\end{equation}
where $f$ is the embedding layer, $v_{q}^{i}$ is the output hidden vector of the QG agent's encoder in the $i$-th iteration, $\mathcal{M}_{q}^{i}$ is the QG agent in $i$-th iteration.

On the decoder side, the agent puts $h_{q}^{i}$, the last layer's hidden state of decoder, into the linear output layer to calculate the word probability distribution with a softmax function:
\begin{align}
    & h_{q}^{i} = \text{Dec}_{\mathcal{M}_{q}^{i}}(v_{q}^{i}) \\
    & \text{P}_{vocab} = softmax(Wh_{q}^{i} + V) 
\end{align}
where $W$ is the set of learnable parameters.


\vspace{0.2cm}
\noindent\textbf{Keyphrase Generation Agent}

Both two generation agents in the multi-task framework has similar structure except 
the input layer. For keyphrase generation agent, $p$ is applied individually to the embedding layer for the embedding matrix $e_k$. Then $e_k$ will be concatenated with $h_{q}^{{i-1}}$ to compose the input of the agent's encoder:
\begin{align}
v_{k}^{i} = \text{Enc}_{\mathcal{M}_{k}^{i}}([h_{q}^{{i-1}}:f(p)])
\end{align}
where $f$ is the embedding layer and $h_q^{i-1}$ is the final hidden state of the \emph{$i - 1$}-th QG agent, $\mathcal{M}_{k}^{i}$ refers to the keyphrase generation agent in $i$-th iteration.


\subsection{Answer Generation}
After the iterative training for \emph{m} epochs, we generate the final answer with the assistance of the optimized question \emph{$q_{m}$} and keyphrase \emph{$k_{m}$}.
We connect $k_m$, $q_m$ and $p$ by a separator \textbf{[CLS]} and input it into the ProphetNet:
\begin{align}
    & v_{a} = \text{Enc}_{\mathcal{M}_{a}}([k_m:q_m:p]) \\
    & h_{a} = \text{Dec}_{\mathcal{M}_{a}}(v_{a}) \\
    & \text{P}_{vocab} = softmax(Wh_{a} + V)
\end{align}
where $\mathcal{M}_{a}$ refers to the Q-K guided answer generation model. 

\section{Experiment}

\begin{table*}[tp]
\centering
\small
\setlength{\tabcolsep}{2mm}{
\begin{tabular}{cccc|ccc}
\hline
\rule{0pt}{5pt} & \multicolumn{3}{c}{Question Generation} & \multicolumn{3}{c}{Answer Generation} \\
\hline
\hline
\rule{0pt}{5pt} Answer-aware models & BLEU-4 & ROUGE-L & METEOR & BLEU4 & ROUGE-L & METEOR \\
\hline
\rule{0pt}{5pt} Seq2Seq & 4.75 & 23.82 & 8.57 & - & - & - \\
\rule{0pt}{5pt} Pointer-Generator & 5.99 & 30.02 & 12.26 & - & - & - \\
\rule{0pt}{5pt} HRED & 6.16 & 32.70 & 12.48 & - & - & - \\
\rule{0pt}{5pt} Transformer & 6.25 & 32.43 & 13.49 & - & - & - \\
\rule{0pt}{5pt} ELMO-QG & 8.23 & 33.26 & 14.35 & - & - & - \\
\rule{0pt}{5pt} AGGCN-QG & \textbf{11.96} & \textbf{34.24} & \textbf{14.94} & - & - & - \\
\hline
\hline
\rule{0pt}{5pt} Question-answer pair generation (our task) & \multicolumn{3}{c}{} & \multicolumn{3}{c}{} \\
\hline
\rule{0pt}{5pt} ProphetNet base & 7.20 & 29.91 & 14.00 & 3.78 & 20.19 & 7.63 \\
\rule{0pt}{5pt} ProphetNet keyphrase guided & 11.18 & 33.29 & 16.01 & 4.57 & 22.01 & 8.15 \\
\rule{0pt}{5pt} Ours(m=1) & 11.18 & 33.29 & 16.01 & 5.18 & 22.86 & 8.34 \\
\rule{0pt}{5pt} Ours(m=2) & \textbf{11.55} & 33.78 & 16.13 & \textbf{6.87} & \textbf{23.41} & \textbf{8.82} \\
\rule{0pt}{5pt} Ours(m=3) & 11.33 & \textbf{33.83} & \textbf{16.22} & 6.12 & 22.91 & 8.38\\
\hline
ProphetNet with golden phrases & 17.84 & 44.67 & 22.35 & - & - & - \\
\rule{0pt}{5pt} ProphetNet answer-aware & 20.53 & 48.52 & 24.52 & - & - & - \\
\hline
\end{tabular}}
\caption{\label{citation-guide} The experiment results on RACE dataset. \emph{m} means the iterative training epochs.}
\label{tab:4}
\end{table*}

\subsection{Experiment Setting}
Our model adopts the transformer-based pre-training model ProphetNet which contains a 12-layer transformer encoder and a 12-layer n-stream self-attention decoder. All of our agents utilize the built-in vocabulary and the tokenization method of BERT \cite{devlin2019bert}. The dimension of the embedding vector is set to 300. The embedding/hidden size is 1024 and the feed-forward filter size is 4096. We use Adam optimizer \cite{kingma2015adam} with a learning rate of 1×$10^{-5}$ and the batch size is set as 10 through the entire training procedure. We train our model on 2 RTX 2080Ti GPUs for about three days.

In the two-stage fine-tuning for keyphrase generation, we set the training epochs as 15 and 10 on SQuAD and RACE respectively. In the later iterative training, we set the training epochs as 15 for QG and 10 for keyphrase generation. 

We carry out the training and inference on EQG-RACE dataset\footnote{https://github.com/jemmryx/EQG-RACE} proposed by \citet{jia2020eqg}. The passage numbers of training set, validation set and test set are respectively 11457, 642, 609. 

We choose BLEU-4, ROUGE and METEOR to evaluate our model’s performance.

\subsection{Comparing Models}
To the best of our knowledge, our work is the first to perform QAG on RACE. For reference, we list some results of answer-aware models that are quoted from \citet{jia2020eqg}.  

\textbf{Seq2Seq} \cite{hosking-2019-evaluating}: A RNN-based seq2seq model with copy machanism.

\textbf{Pointer-generator} \cite{see-etal-2017-get}: A LSTM-based model with pointer machanism.

\textbf{HRED} \cite{gao2019generating}: A seq2seq model with a hierarchical encoder structure to capture both word-level and sentence-level information.

\textbf{Transformer} \cite{vaswani2017attention}: A standard transformer-based Seq2Seq model.

\textbf{ELMo-QG}\cite{zhang2019addressing}: A maxout-pointer model with feature-enriched input.

\textbf{AGGCN-QG} \cite{jia2020eqg}: A gated self-attention maxout-pointer model with a GCN-based encoder to capture the inter-sentences and intra-sentence relations.

For the QAG task, we implemented the following model settings to compare:

\textbf{ProphetNet base}: A basic ProphetNet model to generate question and answer independently, without any extra input information except the passage. 

\textbf{ProphetNet keyphrase guided}: A ProphetNet model to generate question and answer independently, with the guidance of the generated rough keyphrases. 

\textbf{ProphetNet with golden phrases}: A ProphetNet model to generate questions with the guidance of the golden answer phrases, which are constructed by removing the stop words from the ground-truth answers, as discussed in Section 3.2. 

\textbf{ProphetNet answer-aware}: A ProphetNet model to generate questions with the guidance of the ground-truth answers, which can be regarded as the upper bound for QG.

\subsection{Main Results}

The experiment results are shown in Table~\ref{tab:4}. For answer-aware QG, the RNN Seq2Seq model just gets a 4.75 BLEU-4, and the Transformer's performance is also not satisfactory. 

It is exciting to see that our model gets a close performance with the previous state-of-art answer-guided model AGGCN-QG, achieving a 11.55 BLEU-4 and 16.13 METEOR. The answer-agnostic ProphetNet yields a 7.20 BLEU-4 on the QG task and 3.78 on the AG task, demonstrating that even the strong pre-training model can not perform well on this challenging QAG task. Our unified model improves 4.35 points for QG and 3.09 points for AG over the basic ProphetNet model. 


When the iteration epoch \emph{m}=2, we get the best results, but there is no obvious improvement on the results if we continue to increase the number of $m$. Specially, our question-keyphrase iterative agent brings an obvious performance gain on AG.

When we feed the right answer into ProphetNet (ProphetNet answer-aware), we get a quite high performance with a 20.53 BLEU-4, which indicates that our simple method by concatenating the passage $p$ and answer into the input of ProphetNet is effective. When we replace the answer with separate phrases, the performance of QG slightly drops. We will discuss more on this point in Section 5.3.     

\begin{figure*}[t]
\centering 
\subfigure[Shared-encoder]{ 
\label{fig:subfig:a} 
\includegraphics[width=1.9in, height=1.8in]{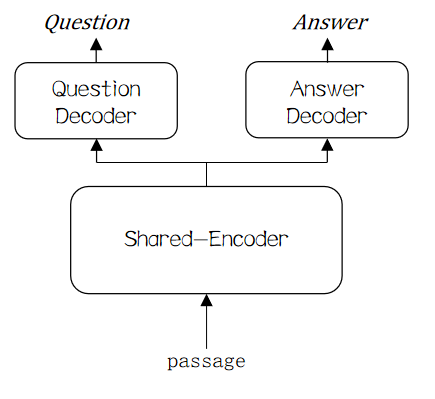}} 
\hspace{1in} 
\subfigure[Q-A iterative]{ 
\label{fig:subfig:b} 
\includegraphics[width=1.9in, height=1.8in]{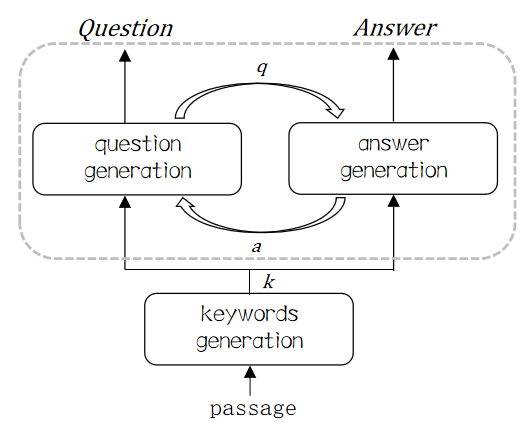}} 
\caption{Two methods for multi-task learning. }
\label{fig:subfig} 
\end{figure*}

\section{Model Analysis}

\subsection{Keyphrase Generation: Mixed Data Augmentation or Two-stage Fine-tuning}
As discussed in Section 3.2, the SQuAD dataset is applied to enhance our training data for rough keyphrase generation. There exist different strategies to exploit the two datasets of SQuAD and RACE. 

\textbf{RACE only}: Only apply RACE data to fine-tune the pre-training model. 

\textbf{SQuAD only}: Only apply SQuAD data to fine-tune the pre-training model. 

\textbf{Mixed data augmentation}: Merge the data from SQuAD and RACE together and fine-tune the pre-training model on the mixed collection. 

\textbf{Two-stage fine-tuning}: Launch a two-stage fine-tuning, first on the larger SQuAD and then on the smaller RACE. 

\begin{table}[htb]
\centering
\begin{tabular}{lllll}
\hline
 & \textbf{B4} & \textbf{R-L} & \textbf{MET} \\
\hline
\hline
ProphetNet base & 7.20 & 29.91 & 14.00 \\
RACE only & 6.84 & 30.46 & 13.72 \\
SQuAD only & 9.73 & 32.94 & 15.96 \\
Mixed datasets & 10.35 & 33.23 & 16.20 \\
Two-stage fine-tuning & 11.18 & 33.29 & 16.01 \\
\hline
\end{tabular}
\caption{\label{citation-guide} The results of generated questions based on different keyphrase generation methods. B4 is BLEU4, R-L is ROUGE-L, and MET is METEOR.}
\label{tab:kg}
\end{table}

We report the experimental results in Table~\ref{tab:kg}. Just applying RACE data for fine-tuning individually even leads to the reduction of the result score, which may be caused by two reasons. First, the data size of RACE is small. Second, adopting the discontinuous answer phrases as the training target may lead to the loss of semantic information, and this is why we introduce SQuAD to enhance our training data. The keyphrases generated by two-stage fine-tuning bring better results than one-stage mixed data.  
Given that the answers of SQuAD are part of the original passage, completely training the keyphrase generation with SQuAD may lead to the degeneration of our model into an extraction model. In contrast, the two-stage fine-tuning can take great advantage of the large scale data of SQuAD as well as avoiding the degeneration mistake.  


\subsection{Multi-task Learning: Shared Encoder or Iterative Training}
We conduct experiments on different multi-task learning methods for Q-A pair joint learning.

\textbf{Shared-encoder architecture}: 
As illustrated in Figure~\ref{fig:subfig:a}, encode passage information with a shared-encoder and generate question and answer with two decoders respectively.

\textbf{Question-answer iterative generation}:
As illustrated in Figure~\ref{fig:subfig:b}, capture keyphrases \emph{k} with a generation agent and iteratively generate question and answer with the input passage \emph{p} and \emph{k}.

\begin{table}[htb]
\centering
\begin{tabular}{lll}
\hline
 & \textbf{Question} & \textbf{Answer} \\
\hline
\hline
ProphetNet base & 7.20 & 3.78 \\
shared-encoder & 3.06 & 1.29 \\
q-a iterative generation & 10.39 & 5.78 \\
q-k iterative generation & 11.55 & 6.87 \\
\hline
\end{tabular}
\caption{\label{citation-guide} The results of  generated questions based on different learning strategies. The metric is BLEU-4.}
\label{tab:ml}
\end{table}

\textbf{Question-keyphrase iterative generation}:
As illustrated in Figure~\ref{fig:qk}, the three stage generation process we adopt in our final model.

We report the BLEU-4 score of the generated Q-A pairs in Table~\ref{tab:ml}. The shared-encoder model just obtains a 3.06 BLEU-4 on question and 1.29 on answer.  It demonstrates that the shared-encoder model can not generate desirable results in our task, and this may be related to the structure of ProphetNet. Iterative training yields obvious performance gain over both the ProphetNet base and shared-encoder method. Specially, 
the Q-K iterative method outperforms the Q-A based one in both QG and AG tasks, because the keyphrases generated in the first stage are relatively rough and should be further optimized. 

\subsection{Key Content: Key Sentences or Key Phrases}
To capture the important content that is worthy of being questioned and answered is vital to our task. 
Aiming at Q-A generation, we can use keyphrases or key sentences to represent the important content of a passage. Key sentences benefit from the complete syntactic and semantic information while keyphrases are more flexible and they will not bring useless information to disturb the generation process. We conduct the following experiments to investigate this issue. 

\textbf{Keyphrase}: Use the rough keyphrase generation agent (discussed in Section 3.2) to obtain the keyphrases from the passage.  

\textbf{Most similar sentence}: Select the key sentence that has the highest matching rate with the keyphrases generated from the rough keyphrase generation agent.

\textbf{Summarized key sentence}:  Apply the pre-trained model for text summarization, BertSum \cite{liu2019fine}, to extract key sentences from the passage. 

The results of these methods can be found in Table~\ref{tab:key}. Overall, the keyphrase based model achieves a better result than ones based on key sentences. In more detail, key sentences extracted by BertSum bring a worse performance, which implies there exactly lies a gap between the existing text summarization and keyphrase extraction tasks orientated towards Q-A generation. On the other side, a slight decline arises on all three metrics when we replace the keyphrases with the most similar sentence, which is probably because some information distortion occurs after the replacement operation.

\begin{table}[htb]
\centering
\begin{tabular}{llll}
\hline
 & \textbf{B4} & \textbf{R-L} & \textbf{MET} \\
\hline
\hline
Generated keyphrases & 11.18 & 33.29 & 16.01 \\
Most similar sentence & 10.49 & 32.27 & 15.39 \\
Extracted key sentence & 8.19 & 29.05 & 13.58 \\
\hline
\end{tabular}
\caption{\label{citation-guide} The result of generated questions based on different key information representation methods}
\label{tab:key}
\end{table}

\begin{table*}[htb]
\small
\centering
\begin{tabular}{llllll}
\hline
 & \textbf{Question Fluency} & \textbf{Answer Fluency} & \textbf{Question Relevance} & \textbf{Answer Relevancy} & \textbf{Answerability} \\
\hline
\hline
ProphetNet & 2.943 & 2.953 & \textbf{2.920} & \textbf{2.927} & 1.689 \\
Our model & \textbf{2.950} & \textbf{2.967} & 2.907 & 2.923 & \textbf{2.640} \\
\hline
Spearman & 0.584 & 0.560 & 0.731 & 0.780 & 0.690 \\
\hline
\end{tabular}
\caption{\label{citation-guide} Human judgement results of generated Q-A pairs. }
\label{tab:7}
\end{table*}

\begin{table*}[htb]
\small
\centering
\begin{tabular}{lp{12.9cm}}
\hline
\textbf{Case 1 --Level 1} & \\
\emph{document} & british food is very different from chinese food . for example , they eat a lot of potatoes. ··· \textcolor{cyan}{they eat a lot of bread with butter for breakfast and usually for one other meal .} ··· \\
\emph{gold QA} & what do they eat when they have bread ? —— butter . \\
\emph{our QA} & what do they eat for breakfast ? —— bread with butter .\\
\hline
\textbf{Case 2 -- Level 2} & \\ 
\emph{document} & take that tiger mom in the ongoing battle between tiger moms , french mamas , and everyone else who wants to know what is the best way to raise their kids , ··· \textcolor{red}{authoritarian parents are more likely to end up with disrespectful children with violent behaviors , the study found ,} compared to parents who listen to their kids with the goal of gaining trust . ··· \textcolor{blue}{to explain the link between parenting style and behavior in kids , the researchers suggested that what matters most is how reasonable kids think their parents ' power is .} ··· \\
\emph{gold QA} & according to the research , what kind of parenting style is likely to cause children 's criminal behaviors ? —— authoritarian parenting .\\
\emph{our QA} & according to the researchers , what matters most about kids ' behavior ? —— how reasonable kids think their parents ' power is . \\
\hline
\textbf{Case 3 -- Level 4} & \\
\emph{document} & have you ever tried broccoli ice cream ? that 's what oliver serves his customers in the new movie oliver 's organic ice cream . ··· \textcolor{blue}{it takes 15 pictures to make just one second of film . to make a movie that lasts one minute , students need to take about 900 frames .} a frame is a picture . ··· \\
\emph{gold QA} & what is the most important thing students learn by making a movie ? —— the way of working with others . \\
\emph{our QA} & how many pictures does it take to make a one - minute film ? —— 15 . \\
\hline
\end{tabular}
\caption{\label{citation-guide} Examples of the Q-A pairs generated by our model. The \textcolor{red}{red} sentences in document refer to the questioned sentences of gold QA, the \textcolor{blue}{blue} ones refer to the questioned sentences of our QA, and the \textcolor{cyan}{cyan} ones are the focus of both reference and the generated results.}
\label{tab:6}
\end{table*}

\section{Human Evaluation}
\subsection{Human Judgement}
Automatic evaluation metrics like BLEU-4 are not well appropriate for the QAG task, so we perform human judgement to compare the ProphetNet base model and our unified model. We randomly selected 100 samples from RACE and asked three annotators to score these generated Q-A pairs in the scale of [1,3], according to five aspects: 

\textbf{Question/Answer Fluency}: which measures whether a question/answer is grammatical and fluent; 

\textbf{Question/Answer Relevancy}: which measures whether the generated question/answer is semantic relevant to the input passage; 

\textbf{Answerability}: which indicates whether the generated question can be answered by the generated answer.

The evaluation results are shown in Table~\ref{tab:7}. The Spearman correlation coefficients between annotators are high. 
Both models achieve nearly full marks on fluency and relevancy due to the powerful performance of pre-training model. Especially, our unified model obtains an obvious improvement on answerability, which demonstrates the effectiveness of our joint learning method. 

\subsection{Case Study}
Further, we make a detailed analysis on the above 100 samples. According to the quality and relevancy with the reference QA, we categorize the generated QA pair into four levels: 

\textbf{level 1}: is of high-quality and similar with references; 

\textbf{level 2}: is of high-quality while has different focus with the reference; 

\textbf{level 3}: has grammatical and syntactic errors; 

\textbf{level 4}: has mismatch error between the generated question and answer. 

We count the proportions of these four levels and display some corresponding case examples in Table~\ref{tab:6}.

We find that 81\% of our results are of high-quality, but most of them (65\%) have different questioned focus with the reference, like Case 2. 
Among them, just 4\% results have grammatical and syntactic errors.
However, our model suffers from the problem of information mismatch when encountering the co-occurrence of complex details in a short piece of text, which causes about 15\% Q-A mismatch error in the generated results. As shown in case 3, according to the passage, the students should make "15 frames" for "1-second film" and "900 frames" for "1-minute film", while our model confuses the correspondence. 


\section{Related Work}
\subsection{Answer-aware Question Generation}



Given the document and answer, Answer-aware Question Generation (AQG) focuses on generating grammatical and answer-related questions, which can help to construct Q-A pairs automatically. Most AQG methods perform sequence-to-sequence generation with neural network models. \citet{song2018leveraging} propose a LSTM-based model with different matching strategies to capture answer-related information among the document. \citet{yao2018teaching} present a Seq2Seq model deployed in GAN with a discriminator, which encourages model to generate more readable and diverse questions with the certain types. \citet{sun2018answer} and \citet{chen2019natural} incorporate lexical features such as POS, NER, and answer position into the encoder for better representing the document. \citet{jia2020eqg} notice the disadvantages when applying Web-extracted data to real-world question generation task and construct a feature-enhanced model on RACE, while it regards answers as available and generates questions with the assistance of answer information.

\subsection{Question-Answer Pair Generation}
Existing QAG methods can be grouped into two categories. First, the joint-learning method conducts question generation and answer extraction simultaneously. \citet{Sachan2018SelfTrainingFJ} propose a self-training method for jointly learning to ask questions as well as answer questions. \citet{tang2017question} regard QA and QG as dual tasks and explicitly leverage their probabilistic correlation to guide the training process of both QA and QG. \citet{wang2017joint} design a generative machine that encodes the document and generates a question (answer) given an answer (question). Second, the pipeline strategy sequentially generates (or extracts) answers and questions with given documents. \citet{du2018harvesting} first identify the question-worthy answer spans from input passage with an extraction model and then generate the answer-aware question. \citet{golub2017two} propose a two-stage SynNet with an answer tagging model and a question synthesis model for question-answer pair generation. \citet{willis2019key} explore two different approaches based on classifier and generative language model respectively to extract high-quality answer phrases from the given passage.


\section{Conclusion}
In this paper, we address the QAG task and propose a unified framework trained on the educational dataset RACE. 
We adopt a three-stage generation method with a rough keyphrase generation model, an iterative message-passing module and a question-keyphrase guided answer generation model. 
Our model achieves close performance as the state-of-the-art answer-aware generation model on QG task, and obtains a great improvement on the answerability of the generated pairs compared to the basic pre-training model. 
There is significant potential for further improvement in our proposed QAG task, to help people produce reading comprehension data in real-world applications.  

\section*{Acknowledgement} 
This work is supported in part by the National
Hi-Tech RD Program of China (No.
2020AAA0106600), the  National  Natural  Science  Foundation  of  China  (No.62076008, No.61773026). 

\bibliographystyle{acl_natbib}

\begin{thebibliography}{36}
\expandafter\ifx\csname natexlab\endcsname\relax\def\natexlab#1{#1}\fi

\bibitem[{Chen et~al.(2019)Chen, Wu, and Zaki}]{chen2019natural}
Yu~Chen, Lingfei Wu, and Mohammed~J Zaki. 2019.
\newblock \href {https://arxiv.org/abs/1910.08832} {Natural question generation
  with reinforcement learning based graph-to-sequence model}.
\newblock \emph{arXiv preprint arXiv:1910.08832}.

\bibitem[{Collobert et~al.(2011)Collobert, Weston, Bottou, Karlen, Kavukcuoglu,
  and Kuksa}]{collobert2011natural}
Ronan Collobert, Jason Weston, L{\'e}on Bottou, Michael Karlen, Koray
  Kavukcuoglu, and Pavel Kuksa. 2011.
\newblock Natural language processing (almost) from scratch.
\newblock \emph{Journal of machine learning research}, 12(ARTICLE):2493--2537.

\bibitem[{Cui et~al.(2021)Cui, Bao, Zu, Guo, Zhao, Zhang, and
  Chen}]{cui2021onestop}
Shaobo Cui, Xintong Bao, Xinxing Zu, Yangyang Guo, Zhongzhou Zhao, Ji~Zhang,
  and Haiqing Chen. 2021.
\newblock Onestop qamaker: Extract question-answer pairs from text in a
  one-stop approach.
\newblock \emph{arXiv e-prints}, pages arXiv--2102.

\bibitem[{Devlin et~al.(2019)Devlin, Chang, Lee, and
  Toutanova}]{devlin2019bert}
Jacob Devlin, Ming-Wei Chang, Kenton Lee, and Kristina Toutanova. 2019.
\newblock \href {https://doi.org/10.18653/v1/N19-1423} {{BERT}: Pre-training of
  deep bidirectional transformers for language understanding}.
\newblock In \emph{Proceedings of the 2019 Conference of the North {A}merican
  Chapter of the Association for Computational Linguistics: Human Language
  Technologies, Volume 1 (Long and Short Papers)}, pages 4171--4186,
  Minneapolis, Minnesota. Association for Computational Linguistics.

\bibitem[{Du and Cardie(2018)}]{du2018harvesting}
Xinya Du and Claire Cardie. 2018.
\newblock \href {https://doi.org/10.18653/v1/P18-1177} {Harvesting
  paragraph-level question-answer pairs from {W}ikipedia}.
\newblock In \emph{Proceedings of the 56th Annual Meeting of the Association
  for Computational Linguistics (Volume 1: Long Papers)}, pages 1907--1917,
  Melbourne, Australia. Association for Computational Linguistics.

\bibitem[{Firat et~al.(2016)Firat, Cho, and Bengio}]{firat2016multi}
Orhan Firat, Kyunghyun Cho, and Yoshua Bengio. 2016.
\newblock \href {https://doi.org/10.18653/v1/N16-1101} {Multi-way, multilingual
  neural machine translation with a shared attention mechanism}.
\newblock In \emph{Proceedings of the 2016 Conference of the North {A}merican
  Chapter of the Association for Computational Linguistics: Human Language
  Technologies}, pages 866--875, San Diego, California. Association for
  Computational Linguistics.

\bibitem[{Gao et~al.(2019)Gao, Bing, Li, King, and Lyu}]{gao2019generating}
Yifan Gao, Lidong Bing, Piji Li, Irwin King, and Michael~R. Lyu. 2019.
\newblock \href {https://doi.org/10.1609/aaai.v33i01.33016423} {Generating
  distractors for reading comprehension questions from real examinations}.
\newblock In \emph{The Thirty-Third {AAAI} Conference on Artificial
  Intelligence, {AAAI} 2019, The Thirty-First Innovative Applications of
  Artificial Intelligence Conference, {IAAI} 2019, The Ninth {AAAI} Symposium
  on Educational Advances in Artificial Intelligence, {EAAI} 2019, Honolulu,
  Hawaii, USA, January 27 - February 1, 2019}, pages 6423--6430. {AAAI} Press.

\bibitem[{Golub et~al.(2017)Golub, Huang, He, and Deng}]{golub2017two}
David Golub, Po-Sen Huang, Xiaodong He, and Li~Deng. 2017.
\newblock \href {https://doi.org/10.18653/v1/D17-1087} {Two-stage synthesis
  networks for transfer learning in machine comprehension}.
\newblock In \emph{Proceedings of the 2017 Conference on Empirical Methods in
  Natural Language Processing}, pages 835--844, Copenhagen, Denmark.
  Association for Computational Linguistics.

\bibitem[{Heilman(2011)}]{heilman2011automatic}
Michael Heilman. 2011.
\newblock Automatic factual question generation from text.
\newblock \emph{Language Technologies Institute School of Computer Science
  Carnegie Mellon University}, 195.

\bibitem[{Hochreiter and Schmidhuber(1997)}]{hochreiter1997long}
Sepp Hochreiter and J{\"u}rgen Schmidhuber. 1997.
\newblock Long short-term memory.
\newblock \emph{Neural computation}, 9(8):1735--1780.

\bibitem[{Hosking and Riedel(2019)}]{hosking-2019-evaluating}
Tom Hosking and Sebastian Riedel. 2019.
\newblock \href {https://doi.org/10.18653/v1/N19-1237} {Evaluating rewards for
  question generation models}.
\newblock In \emph{Proceedings of the 2019 Conference of the North {A}merican
  Chapter of the Association for Computational Linguistics: Human Language
  Technologies, Volume 1 (Long and Short Papers)}, pages 2278--2283,
  Minneapolis, Minnesota. Association for Computational Linguistics.

\bibitem[{Jia et~al.(2020)Jia, Zhou, Sun, and Wu}]{jia2020eqg}
Xin Jia, Wenjie Zhou, Xu~Sun, and Yunfang Wu. 2020.
\newblock \href {https://arxiv.org/abs/2012.06106} {Eqg-race: Examination-type
  question generation}.
\newblock \emph{arXiv preprint arXiv:2012.06106}.

\bibitem[{Kingma and Ba(2015)}]{kingma2015adam}
Diederik~P. Kingma and Jimmy Ba. 2015.
\newblock \href {http://arxiv.org/abs/1412.6980} {Adam: {A} method for
  stochastic optimization}.
\newblock In \emph{3rd International Conference on Learning Representations,
  {ICLR} 2015, San Diego, CA, USA, May 7-9, 2015, Conference Track
  Proceedings}.

\bibitem[{Krishna and Iyyer(2019)}]{krishna2019generating}
Kalpesh Krishna and Mohit Iyyer. 2019.
\newblock \href {https://doi.org/10.18653/v1/P19-1224} {Generating
  question-answer hierarchies}.
\newblock In \emph{Proceedings of the 57th Annual Meeting of the Association
  for Computational Linguistics}, pages 2321--2334, Florence, Italy.
  Association for Computational Linguistics.

\bibitem[{Lai et~al.(2017)Lai, Xie, Liu, Yang, and Hovy}]{lai2017race}
Guokun Lai, Qizhe Xie, Hanxiao Liu, Yiming Yang, and Eduard Hovy. 2017.
\newblock \href {https://doi.org/10.18653/v1/D17-1082} {{RACE}: Large-scale
  {R}e{A}ding comprehension dataset from examinations}.
\newblock In \emph{Proceedings of the 2017 Conference on Empirical Methods in
  Natural Language Processing}, pages 785--794, Copenhagen, Denmark.
  Association for Computational Linguistics.

\bibitem[{Lewis et~al.(2020)Lewis, Liu, Goyal, Ghazvininejad, Mohamed, Levy,
  Stoyanov, and Zettlemoyer}]{lewis2020bart}
Mike Lewis, Yinhan Liu, Naman Goyal, Marjan Ghazvininejad, Abdelrahman Mohamed,
  Omer Levy, Veselin Stoyanov, and Luke Zettlemoyer. 2020.
\newblock \href {https://doi.org/10.18653/v1/2020.acl-main.703} {{BART}:
  Denoising sequence-to-sequence pre-training for natural language generation,
  translation, and comprehension}.
\newblock In \emph{Proceedings of the 58th Annual Meeting of the Association
  for Computational Linguistics}, pages 7871--7880, Online. Association for
  Computational Linguistics.

\bibitem[{Liu et~al.(2020{\natexlab{a}})Liu, Wei, Niu, Chen, and
  He}]{liu2020asking}
Bang Liu, Haojie Wei, Di~Niu, Haolan Chen, and Yancheng He. 2020{\natexlab{a}}.
\newblock \href {https://doi.org/10.1145/3366423.3380270} {Asking questions the
  human way: Scalable question-answer generation from text corpus}.
\newblock In \emph{{WWW} '20: The Web Conference 2020, Taipei, Taiwan, April
  20-24, 2020}, pages 2032--2043. {ACM} / {IW3C2}.

\bibitem[{Liu et~al.(2020{\natexlab{b}})Liu, Gong, Fu, Yan, Chen, Lv, Duan, and
  Zhou}]{liu2020tell}
Dayiheng Liu, Yeyun Gong, Jie Fu, Yu~Yan, Jiusheng Chen, Jiancheng Lv, Nan
  Duan, and Ming Zhou. 2020{\natexlab{b}}.
\newblock \href {https://doi.org/10.18653/v1/2020.emnlp-main.467} {Tell me how
  to ask again: Question data augmentation with controllable rewriting in
  continuous space}.
\newblock In \emph{Proceedings of the 2020 Conference on Empirical Methods in
  Natural Language Processing (EMNLP)}, pages 5798--5810, Online. Association
  for Computational Linguistics.

\bibitem[{Liu et~al.(2021)Liu, Yan, Gong, Qi, Zhang, Jiao, Chen, Fu, Shou,
  Gong, Wang, Chen, Jiang, Lv, Zhang, Wu, Zhou, and Duan}]{liu2020glge}
Dayiheng Liu, Yu~Yan, Yeyun Gong, Weizhen Qi, Hang Zhang, Jian Jiao, Weizhu
  Chen, Jie Fu, Linjun Shou, Ming Gong, Pengcheng Wang, Jiusheng Chen, Daxin
  Jiang, Jiancheng Lv, Ruofei Zhang, Winnie Wu, Ming Zhou, and Nan Duan. 2021.
\newblock \href {https://doi.org/10.18653/v1/2021.findings-acl.36} {{GLGE}: A
  new general language generation evaluation benchmark}.
\newblock In \emph{Findings of the Association for Computational Linguistics:
  ACL-IJCNLP 2021}, pages 408--420, Online. Association for Computational
  Linguistics.

\bibitem[{Liu(2019)}]{liu2019fine}
Yang Liu. 2019.
\newblock \href {https://arxiv.org/abs/1903.10318} {Fine-tune bert for
  extractive summarization}.
\newblock \emph{arXiv preprint arXiv:1903.10318}.

\bibitem[{Nguyen et~al.(2016)Nguyen, Rosenberg, Song, Gao, Tiwary, Majumder,
  and Deng}]{nguyen2016ms}
Tri Nguyen, Mir Rosenberg, Xia Song, Jianfeng Gao, Saurabh Tiwary, Rangan
  Majumder, and Li~Deng. 2016.
\newblock Ms marco: A human generated machine reading comprehension dataset.
\newblock In \emph{CoCo@ NIPS}.

\bibitem[{Qi et~al.(2020)Qi, Yan, Gong, Liu, Duan, Chen, Zhang, and
  Zhou}]{qi2020prophetnet}
Weizhen Qi, Yu~Yan, Yeyun Gong, Dayiheng Liu, Nan Duan, Jiusheng Chen, Ruofei
  Zhang, and Ming Zhou. 2020.
\newblock \href {https://doi.org/10.18653/v1/2020.findings-emnlp.217}
  {{P}rophet{N}et: Predicting future n-gram for
  sequence-to-{S}equence{P}re-training}.
\newblock In \emph{Findings of the Association for Computational Linguistics:
  EMNLP 2020}, pages 2401--2410, Online. Association for Computational
  Linguistics.

\bibitem[{Rajpurkar et~al.(2016)Rajpurkar, Zhang, Lopyrev, and
  Liang}]{rajpurkar2016squad}
Pranav Rajpurkar, Jian Zhang, Konstantin Lopyrev, and Percy Liang. 2016.
\newblock \href {https://doi.org/10.18653/v1/D16-1264} {{SQ}u{AD}: 100,000+
  questions for machine comprehension of text}.
\newblock In \emph{Proceedings of the 2016 Conference on Empirical Methods in
  Natural Language Processing}, pages 2383--2392, Austin, Texas. Association
  for Computational Linguistics.

\bibitem[{Sachan and Xing(2018)}]{Sachan2018SelfTrainingFJ}
Mrinmaya Sachan and Eric Xing. 2018.
\newblock \href {https://doi.org/10.18653/v1/N18-1058} {Self-training for
  jointly learning to ask and answer questions}.
\newblock In \emph{Proceedings of the 2018 Conference of the North {A}merican
  Chapter of the Association for Computational Linguistics: Human Language
  Technologies, Volume 1 (Long Papers)}, pages 629--640, New Orleans,
  Louisiana. Association for Computational Linguistics.

\bibitem[{See et~al.(2017)See, Liu, and Manning}]{see-etal-2017-get}
Abigail See, Peter~J. Liu, and Christopher~D. Manning. 2017.
\newblock \href {https://doi.org/10.18653/v1/P17-1099} {Get to the point:
  Summarization with pointer-generator networks}.
\newblock In \emph{Proceedings of the 55th Annual Meeting of the Association
  for Computational Linguistics (Volume 1: Long Papers)}, pages 1073--1083,
  Vancouver, Canada. Association for Computational Linguistics.

\bibitem[{Song et~al.(2018)Song, Wang, Hamza, Zhang, and
  Gildea}]{song2018leveraging}
Linfeng Song, Zhiguo Wang, Wael Hamza, Yue Zhang, and Daniel Gildea. 2018.
\newblock \href {https://doi.org/10.18653/v1/N18-2090} {Leveraging context
  information for natural question generation}.
\newblock In \emph{Proceedings of the 2018 Conference of the North {A}merican
  Chapter of the Association for Computational Linguistics: Human Language
  Technologies, Volume 2 (Short Papers)}, pages 569--574, New Orleans,
  Louisiana. Association for Computational Linguistics.

\bibitem[{Sun et~al.(2018)Sun, Liu, Lyu, He, Ma, and Wang}]{sun2018answer}
Xingwu Sun, Jing Liu, Yajuan Lyu, Wei He, Yanjun Ma, and Shi Wang. 2018.
\newblock \href {https://doi.org/10.18653/v1/D18-1427} {Answer-focused and
  position-aware neural question generation}.
\newblock In \emph{Proceedings of the 2018 Conference on Empirical Methods in
  Natural Language Processing}, pages 3930--3939, Brussels, Belgium.
  Association for Computational Linguistics.

\bibitem[{Tang et~al.(2017)Tang, Duan, Qin, Yan, and Zhou}]{tang2017question}
Duyu Tang, Nan Duan, Tao Qin, Zhao Yan, and Ming Zhou. 2017.
\newblock \href {https://arxiv.org/abs/1706.02027} {Question answering and
  question generation as dual tasks}.
\newblock \emph{arXiv preprint arXiv:1706.02027}.

\bibitem[{Trischler et~al.(2017)Trischler, Wang, Yuan, Harris, Sordoni,
  Bachman, and Suleman}]{trischler2017newsqa}
Adam Trischler, Tong Wang, Xingdi Yuan, Justin Harris, Alessandro Sordoni,
  Philip Bachman, and Kaheer Suleman. 2017.
\newblock \href {https://doi.org/10.18653/v1/W17-2623} {{N}ews{QA}: A machine
  comprehension dataset}.
\newblock In \emph{Proceedings of the 2nd Workshop on Representation Learning
  for {NLP}}, pages 191--200, Vancouver, Canada. Association for Computational
  Linguistics.

\bibitem[{Vaswani et~al.(2017)Vaswani, Shazeer, Parmar, Uszkoreit, Jones,
  Gomez, Kaiser, and Polosukhin}]{vaswani2017attention}
Ashish Vaswani, Noam Shazeer, Niki Parmar, Jakob Uszkoreit, Llion Jones,
  Aidan~N. Gomez, Lukasz Kaiser, and Illia Polosukhin. 2017.
\newblock \href
  {https://proceedings.neurips.cc/paper/2017/hash/3f5ee243547dee91fbd053c1c4a845aa-Abstract.html}
  {Attention is all you need}.
\newblock In \emph{Advances in Neural Information Processing Systems 30: Annual
  Conference on Neural Information Processing Systems 2017, December 4-9, 2017,
  Long Beach, CA, {USA}}, pages 5998--6008.

\bibitem[{Wagner and Bolloju(2005)}]{wagner2005supporting}
Christian Wagner and Narasimha Bolloju. 2005.
\newblock Supporting knowledge management in organizations with conversational
  technologies: Discussion forums, weblogs, and wikis.
\newblock \emph{Journal of Database Management}, 16(2):I.

\bibitem[{Wang et~al.(2017)Wang, Yuan, and Trischler}]{wang2017joint}
Tong Wang, Xingdi Yuan, and Adam Trischler. 2017.
\newblock \href {https://arxiv.org/abs/1706.01450} {A joint model for question
  answering and question generation}.
\newblock \emph{arXiv preprint arXiv:1706.01450}.

\bibitem[{Willis et~al.(2019)Willis, Davis, Ruan, Manoharan, Landay, and
  Brunskill}]{willis2019key}
Angelica Willis, Glenn Davis, Sherry Ruan, Lakshmi Manoharan, James Landay, and
  Emma Brunskill. 2019.
\newblock Key phrase extraction for generating educational question-answer
  pairs.
\newblock In \emph{Proceedings of the Sixth (2019) ACM Conference on Learning@
  Scale}, pages 1--10.

\bibitem[{Yang et~al.(2019)Yang, Dai, Yang, Carbonell, Salakhutdinov, and
  Le}]{yang2019xlnet}
Zhilin Yang, Zihang Dai, Yiming Yang, Jaime~G. Carbonell, Ruslan Salakhutdinov,
  and Quoc~V. Le. 2019.
\newblock \href
  {https://proceedings.neurips.cc/paper/2019/hash/dc6a7e655d7e5840e66733e9ee67cc69-Abstract.html}
  {Xlnet: Generalized autoregressive pretraining for language understanding}.
\newblock In \emph{Advances in Neural Information Processing Systems 32: Annual
  Conference on Neural Information Processing Systems 2019, NeurIPS 2019,
  December 8-14, 2019, Vancouver, BC, Canada}, pages 5754--5764.

\bibitem[{Yao et~al.(2018)Yao, Zhang, Luo, Tao, and Wu}]{yao2018teaching}
Kaichun Yao, Libo Zhang, Tiejian Luo, Lili Tao, and Yanjun Wu. 2018.
\newblock \href {https://doi.org/10.24963/ijcai.2018/632} {Teaching machines to
  ask questions}.
\newblock In \emph{Proceedings of the Twenty-Seventh International Joint
  Conference on Artificial Intelligence, {IJCAI} 2018, July 13-19, 2018,
  Stockholm, Sweden}, pages 4546--4552. ijcai.org.

\bibitem[{Zhang and Bansal(2019)}]{zhang2019addressing}
Shiyue Zhang and Mohit Bansal. 2019.
\newblock \href {https://doi.org/10.18653/v1/D19-1253} {Addressing semantic
  drift in question generation for semi-supervised question answering}.
\newblock In \emph{Proceedings of the 2019 Conference on Empirical Methods in
  Natural Language Processing and the 9th International Joint Conference on
  Natural Language Processing (EMNLP-IJCNLP)}, pages 2495--2509, Hong Kong,
  China. Association for Computational Linguistics.

\end{thebibliography}

\end{document}